\documentclass[10pt,twocolumn,letterpaper]{article}

\usepackage{iccv}
\usepackage{times}
\usepackage{epsfig}
\usepackage{graphicx}
\usepackage{animate}
\usepackage{amsmath}
\usepackage{amssymb}
\usepackage{booktabs}
\usepackage{float}
\usepackage{url}
\usepackage{enumitem}
\usepackage{multirow}
\usepackage{algorithm}
\usepackage{algorithmic}
\usepackage{xcolor}
\usepackage{listings}

\setlist{leftmargin=*}

\newcommand{\authorskip}{\hspace{2.5mm}}

\usepackage[pagebackref=true,breaklinks=true,letterpaper=true,colorlinks,bookmarks=false]{hyperref}

\iccvfinalcopy

\ificcvfinal\fi

\begin{document}

\title{All-to-key Attention for Arbitrary Style Transfer}

\author{
 Mingrui Zhu$^{*1}$ \authorskip Xiao He$^{*1}$ \authorskip Nannan Wang$^{\dagger1}$ \authorskip Xiaoyu Wang$^{2}$ \authorskip Xinbo Gao$^{3}$ \\[2mm]
 \small $^{*}$equal technical contribution \qquad $^{\dagger}$corresponding author \\[2mm]
 \small $^{1}$State Key Laboratory of Integrated Services Networks, Xidian University, Xi’an, China \\[2mm]
 \small $^{2}$School of Computer Science and Technology, University of Science and Technology of China, Hefei, China \\[2mm]
 \small $^{3}$Chongqing Key Laboratory of Image Cognition, Chongqing University of Posts and Telecommunications, Chongqing, China\vspace{-4mm} 
}

\maketitle

\ificcvfinal\thispagestyle{empty}\fi

\begin{abstract}
   Attention-based arbitrary style transfer studies have shown promising performance in synthesizing vivid local style details. They typically use the all-to-all attention mechanism---each position of content features is fully matched to all positions of style features. However, all-to-all attention tends to generate distorted style patterns and has quadratic complexity, limiting the effectiveness and efficiency of arbitrary style transfer. In this paper, we propose a novel all-to-key attention mechanism---each position of content features is matched to stable key positions of style features---that is more in line with the characteristics of style transfer. Specifically, it integrates two newly proposed attention forms: distributed and progressive attention. Distributed attention assigns attention to key style representations that depict the style distribution of local regions; Progressive attention pays attention from coarse-grained regions to fine-grained key positions. The resultant module, dubbed StyA2K, shows extraordinary performance in preserving the semantic structure and rendering consistent style patterns. Qualitative and quantitative comparisons with state-of-the-art methods demonstrate the superior performance of our approach.
\end{abstract}\vspace{-4mm}

\section{Introduction}
\label{sec:introduction}

Arbitrary style transfer (AST) is an important computer vision task. It aims to render a natural image (i.e., content image) with the artistic style of an arbitrary painting (i.e., style image), enabling the generated image to imitate any artistic style. There have been notable improvements in feature transformation modules \cite{huang2017arbitrary, sheng2018avatar, gu2018arbitrary, park2019arbitrary, zhang2019multimodal, deng2020arbitrary, liu2021adaattn, kalischek2021light, zhang2022exact}, novel architectures \cite{li2017universal, shen2018neural, an2021artflow, wu2021styleformer, deng2022stytr2}, and practical objectives \cite{kotovenko2019disentanglement, cheng2021style, zhang2022domain}. The core of AST is the matching of content features and style features in the feed-forward procedure. Holistic feature distribution matching \cite{huang2017arbitrary, li2017universal, kalischek2021light, zhang2022exact} and locality-aware feature matching \cite{sheng2018avatar, gu2018arbitrary, park2019arbitrary, zhang2019multimodal, liu2021adaattn} are two categories of existing approaches. 

\begin{figure}[t]
\centering
\begin{minipage}{0.98\linewidth}
   \animategraphics[width=0.32\linewidth, autoplay, loop]{15}{fig/video_compressed/input/}{1}{50}
   \animategraphics[width=0.32\linewidth, autoplay, loop]{15}{fig/video_compressed/adaattn/}{1}{50}
   \animategraphics[width=0.32\linewidth, autoplay, loop]{15}{fig/video_compressed/stya2k/}{1}{50}
\end{minipage} \\
\begin{minipage}{0.98\linewidth}
   \includegraphics[width=0.32\linewidth]{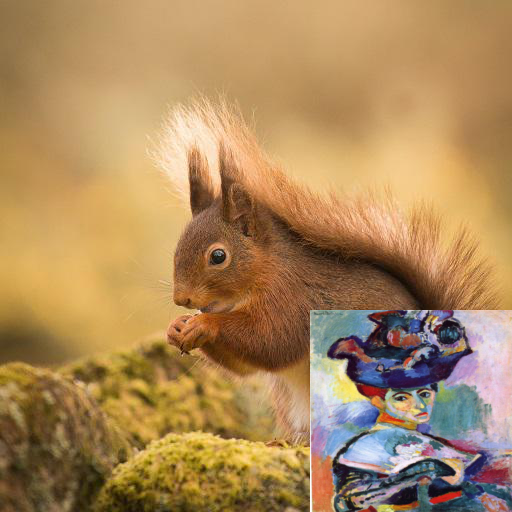}
   \includegraphics[width=0.32\linewidth]{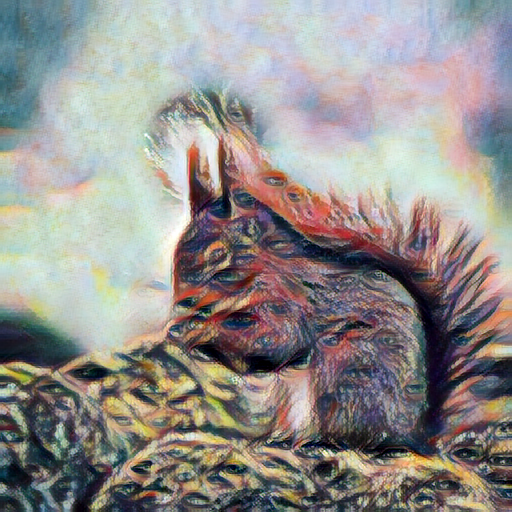}
   \includegraphics[width=0.32\linewidth]{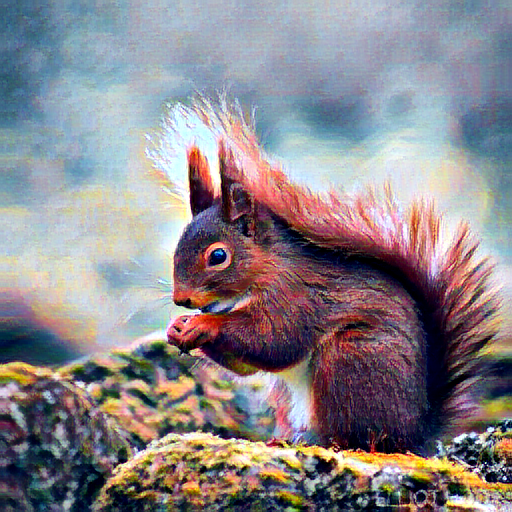}
\end{minipage}
\caption{Image/Video style transfer results of AdaAttN \cite{liu2021adaattn} (second column) and our StyA2K (third column). Adobe Acrobat is recommended to view the videos in the first row.}
\vspace{-2mm}
\label{fig:intro_advan}
\end{figure}

The attention-based method is the research focus of the locality-aware feature matching category for its capability to capture long-range dependencies. Typically, it establishes a dense correspondence between point-wise tokens of the content and style features via an all-to-all attention mechanism \cite{zhao2021improved}. The transferred feature of each local position is computed as the weighted sum of the local style features of all positions, where the weights are computed by applying a $softmax$ function to the normalized dot products’ results. Despite its encouraging results, the attention-based AST method suffers from two main predicaments. The first one is the introduction of distorted style patterns and unstable matching effects. As shown in Fig. \ref{fig:intro_advan}, the image stylization result of AdaAttN is seriously affected by eye patterns, which significantly affects visual perception. Besides, the video stylization result of AdaAttN has an apparent flickering phenomenon, which also affects the visual quality. The second one is the high computational complexity. Handling image features with high spatial resolution and multiple layers with all-to-all attention requires significant computational consumption.

We argue that the inherent limitations of the all-to-all attention mechanism cause the above problems. (1) All-to-all attention has no error tolerance and is sensitive to position variation. It tends to concentrate on the most similar value since $softmax$ shows strong exclusiveness in attention score due to exponential computation. A distorted style pattern appears when the most similar key is semantically distinct from the query. For example, the reason for the appearance of irrational eye patterns in AdaAttN's results is that the all-to-all attention almost exclusively concentrates on the eye patterns in the style image for the positions with edge-like patterns in the content image. Its sensitivity is embodied in that slight changes in query at different positions will lead to complete semantic changes in matched keys. For instance, the object motion and the light change in the video cause severe flickering phenomena between consecutive stylized frames that are frame-by-frame independently generated by AdaAttN. (2) All-to-all attention has quadratic computational complexity since it establishes a fully connected correspondence between queries and keys. Its computational complexity scales quadratically to image size.

\begin{figure}[t]
\centering
\includegraphics[width=0.80\linewidth]{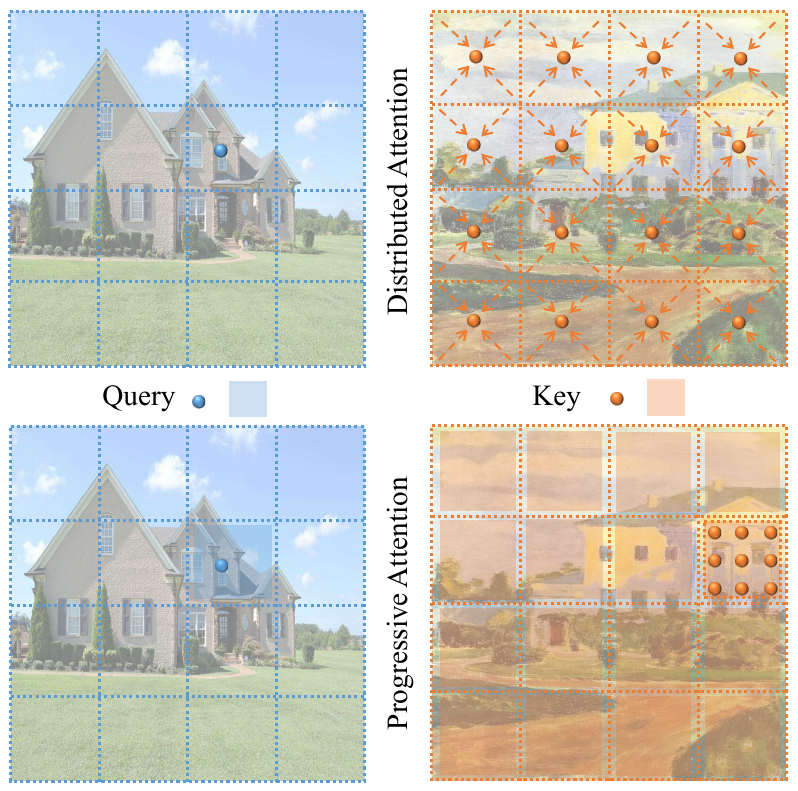}
\caption{Illustration of A2K attention.}
\vspace{-4mm}
\label{fig:intro_dp}
\end{figure}

How to maintain the advantage of the attention mechanism (enhancing local semantics) and mitigate the disadvantages of all-to-all attention (no error tolerance, unstable, time-consuming)? Our solution is a novel all-to-key attention (A2K) mechanism that matches each query with stable ``key'' keys. A2k comprises two inventions. (1) It learns distributed keys that depict the style distribution of all local regions of the style features; thus, each query of the content features is matched with these stable and representative keys. (2) It gradually concentrates its attention from coarse-grained regions to fine-grained keys; thus, queries within a local region are matched to the same stable keys within a local region. The former, dubbed as distributed attention (DA), improves matching error tolerance since the most similar key matched is a regional style representation rather than an isolated position. The latter, dubbed as progressive attention (PA), mitigates the problem of ``cannot see the woods for the trees'' existing in all-to-all attention and ensures semantic correctness from a more macro perspective. In addition, the keys of DA and PA are stable, so A2K can reduce the sensitivity of position variation and render consistent style patterns. Furthermore, we implement DA and PA in a blocked, sparse fashion, saving considerable computational costs. Finally, an effective and efficient arbitrary \textbf{sty}le transfer model based on \textbf{a}ll-\textbf{t}o-\textbf{k}ey attention (StyA2K) arrives. We summarize the main contributions of this paper as the following points:

\begin{itemize}
\item We point out the inherent limitations of all-to-all attention and present a novel all-to-key attention mechanism for effective and efficient arbitrary style transfer.
\item We propose distributed attention to improve matching error tolerance and progressive attention to ensure semantic correctness, both of which enjoy the stable matching property and save significant computational consumption.
\item We conduct extensive experiments to demonstrate the superiority of our approach over state-of-the-art methods in preserving semantic structures and rendering consistent style patterns.
\end{itemize}

\section{Related Work}
\label{sec:related_work}

\subsection{Arbitrary Style Transfer}

Neural style transfer has become a hot-spot topic and has attracted wide attention since the pioneering work of Gatys \etal~\cite{gatys2016image}. Follow-up studies \cite{johnson2016perceptual, li2016combining, li2016precomputed, ulyanov2016texture, chen2017stylebank, liao2017visual, li2017diversified, dumoulin2016learned, gatys2017controlling, ulyanov2017improved, wang2017multimodal} are devoted to improving the capabilities of neural style transfer algorithms in terms of visual quality, computational efficiency, style diversity, structural consistency, and factor controllability. Arbitrary style transfer \cite{chen2016fast, huang2017arbitrary, li2017universal, sheng2018avatar, gu2018arbitrary, li2019learning, park2019arbitrary, yao2019attention, zhang2019multimodal, kotovenko2019disentanglement, jing2020dynamic, deng2020arbitrary, kalischek2021light, liu2021adaattn, an2021artflow, wu2021styleformer, cheng2021style, deng2022stytr2, zhang2022exact, zhang2022domain} has received increasing attention recently, depending on its advantage of using a single feed-forward neural model to transfer the style of an arbitrary image. Existing AST methods can be divided into two main categories: holistic feature distribution matching method and locality-aware feature matching method.

The holistic feature distribution matching method adjusts the holistic content feature distribution to match the style feature distribution. Based on the Gaussian prior assumption, AdaIN \cite{huang2017arbitrary} and WCT \cite{li2017universal} match feature distributions with first- or second-order statistics. The method introduced in \cite{li2019learning} makes the transformation matrix learnable. In order to break through the theoretical and practical limitations of first-order and second-order statistics, high-order statistics are introduced in \cite{kalischek2021light} and \cite{zhang2022exact} to perform more exact distribution matching. 

By comparison, the locality-aware feature matching method emphasizes the consistency of local semantics when matching content and style features. StyleSwap \cite{chen2016fast} replaces content features patch-by-patch with the style features. The deep feature reshuffle module introduced in \cite{gu2018arbitrary} reshuffles the style features according to the content features via a constrained normalized cross-correlation. Avatar-Net \cite{sheng2018avatar} decorates the content features with aligned style features obtained through a relaxed normalized cross-correlation. MST \cite{zhang2019multimodal} employs clustering to divide style features into multi-modal style representations, which are matched with local content features via graph-based style matching. With the rise of self-attention \cite{vaswani2017attention}, many studies \cite{park2019arbitrary, yao2019attention, deng2020arbitrary, liu2021adaattn} apply it to AST. SANet \cite{park2019arbitrary} directly adopts attention-based feature matching for AST. AdaAttN \cite{liu2021adaattn} adopts attention scores to adaptively perform attentive normalization on the content features by calculating the per-point attention-weighted mean and variance of style features. However, these methods neglect the limitations of all-to-all attention and inevitably produce compromised results. 

\subsection{Attention Mechanism}

Since being proposed, the attention mechanism has been widely used in the field of natural language processing (NLP) \cite{vaswani2017attention, yang2019xlnet} and computer vision \cite{wang2018non, dosovitskiy2020image}. With the development of the Transformer \cite{liu2021swin, vaswani2021scaling, zhao2021improved, yu2022boat, zhao2021improved}, extensive variants of attention mechanisms are introduced to improve its capability and computational efficiency. Discovering the scalability and availability of attention mechanism in computing dense matching, many studies \cite{zhang2020cross, jiang2020psgan, park2019arbitrary, liu2021adaattn} adapt self-attention to match features with distinct distributions. Due to the distinctions in content and style, directly applying all-to-all attention to their matching causes many problems. This work fully considers content-style matching characteristics and explores a novel attention mechanism more suitable for AST tasks.

\section{Method}
\label{sec:method}

\subsection{Overall Architecture}

\begin{figure}[t]
\centering
\includegraphics[width=0.92\linewidth]{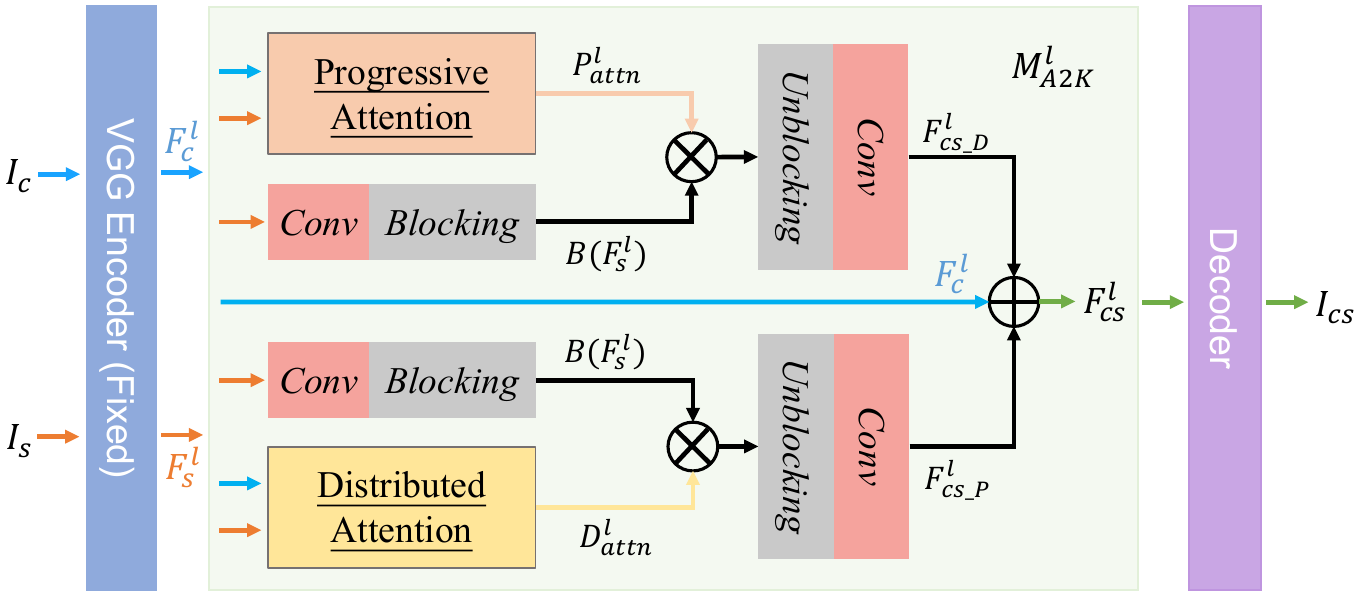}
\caption{Framework of StyA2K.}
\label{fig:meth_arch}
\end{figure}

An overview of our framework is presented in Fig. \ref{fig:meth_arch}. Given a content image $I_c$ and a style image $I_s$, a pre-trained VGG-19 \cite{simonyan2014very} network with fixed parameters is utilized as an encoder $E_{nc}$ to extract their multi-scale features. Extracted features at each layer $l$ can be denoted as:
\begin{equation}
F_c^l = E_{nc}(I_c), F_s^l = E_{nc}(I_s), 
\end{equation}
where $l \in ReLU \{3\_1, 4\_1, 5\_1$\}, ${F_*^l}\in\mathfrak{R}^{C_l \times H_l \times W_l}$ and $*$ can be $c$ or $s$ representing content and style respectively. 

\begin{figure*}[t]
\centering
\includegraphics[width=0.96\linewidth]{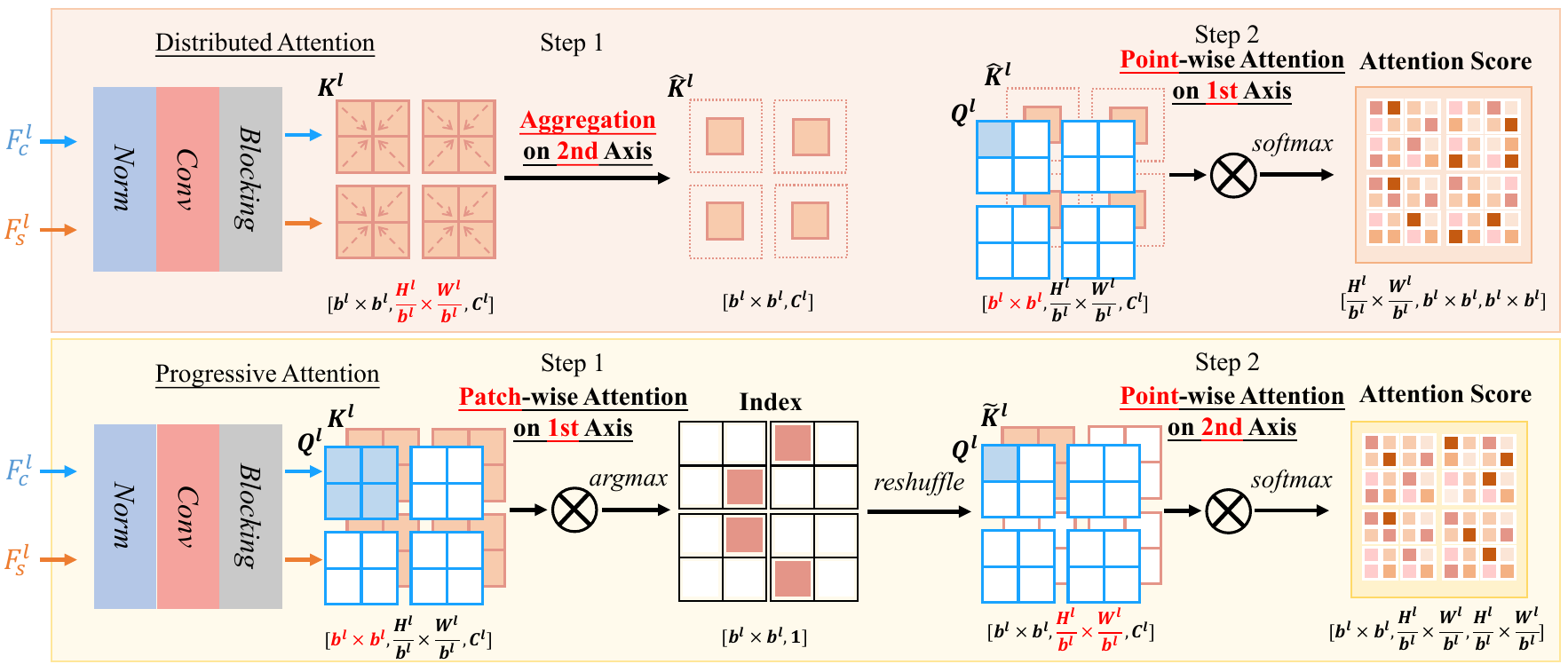}
\caption{Distributed attention (DA) and progressive attention (PA) are implemented in a blocked, sparse fashion. DA aggregates regional style information to obtain representative keys $\widehat{K}^l$ in the first step and then calculates point-wise attention score along the first (global) axis of $Q^l$ and $\widehat{K}^l$ in the second step. PA calculates patch-wise similarity index along the first axis of $Q^l$ and $K^l$ in the first step to reshuffle $K^l$ to $\widetilde{K}^l$ and then calculates the point-wise attention score on the second (regional) axis of $Q^l$ and $\widetilde{K}^l$ in the second step.}
\label{fig:meth_dp}
\end{figure*}

The key ingredient of this framework is the A2K module. It integrates two effective and efficient attention forms (distributed attention and progressive attention as illustrated in Fig. \ref{fig:meth_dp}) to establish meaningful sparse correspondence between the content feature $F_c^l$ and the style feature $F_s^l$ and thus faithfully synthesize the transferred features $F_{cs}^l$: 
\begin{equation}
F_{cs}^l = M_{A2K}^l(F_c^l, F_s^l),
\end{equation}
where $M_{A2K}^l$ denotes the A2K module and $l$ indicates that $M_{A2K}^l$ operates on each layer of the multi-scale features. 

Finally we can invert the multi-scale transferred features $\{F_{cs}^l\}$ to the stylized image $I_{cs}$ through a decoder $D_{ec}$: 
\begin{equation}
I_{cs} = D_{ec}(\{F_{cs}^l\}).
\end{equation}
The decoder implemented in this work follows the setting of \cite{liu2021adaattn}, which mirrors the encoder and takes the multi-scale transferred features as input.

\subsection{All-to-key Attention}

\begin{figure}[t]
\centering
\includegraphics[width=0.85\linewidth]{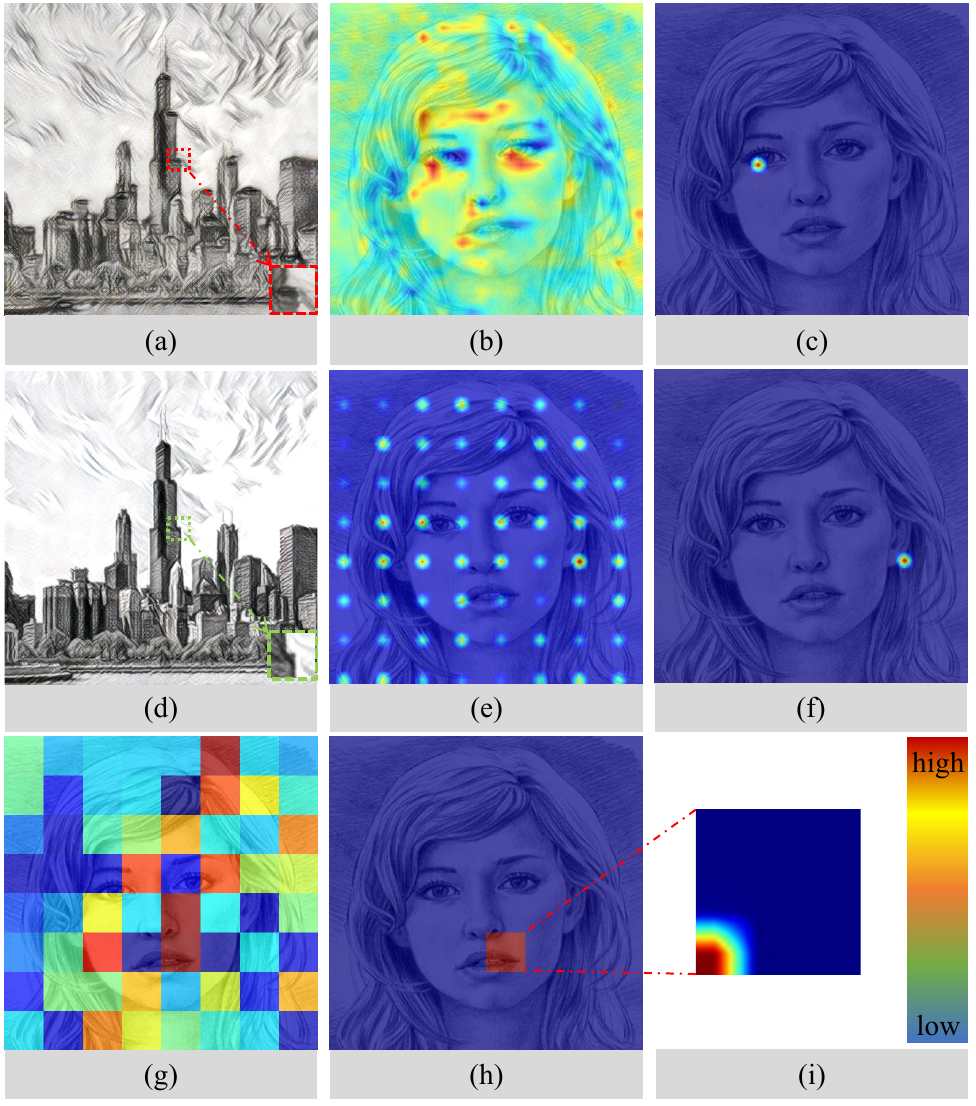}
\caption{Visualization of attention distribution. (a) Stylized result of AdaAttN. (b) and (c) All-to-all attention distribution before and after $softmax$. (d) Stylized result of StyA2K. (e), (f) Distributed attention distribution before and after $softmax$. (g), (h) Progressive attention distribution on the coarse-grained region before and after $argmax$. (i) Progressive attention distribution on the fine-grained position after $softmax$.}
\vspace{-2mm}
\label{fig:meth_vis}
\end{figure}

\textbf{Revisit All-to-all Attention in AST.} Attention mechanism in AST is derived from self-attention \cite{vaswani2017attention} and is commonly used as a locality-aware feature matching method. Formally, given the content feature $F_c^l$ and the style feature $F_s^l$ with the size of $(H^l, W^l, C^l)$ at layer $l$, $Q$ (query), $K$ (key) and $V$ (value) are formulated as:
\begin{equation}
\begin{aligned}
&Q = f(N(F_c^l)), K = g(N(F_s^l)), V = h(F_c^l),
\end{aligned}
\end{equation}
where $f$, $g$, and $h$ are learnable convolution layers and $N(*)$ denotes normalization operation. The attention score $AttN$ can be calculated as:
\begin{equation}
AttN = softmax(Q \cdot K^\mathrm{T}),
\end{equation}
where $\cdot$ denotes the dot product. The attention score $AttN$, with the size of $(H \times W, H \times W)$, is the dense similarity correspondence matrix of all the tokens in $F_c^l$ and $F_s^l$. Since this attention mechanism regards feature vectors of all spatial positions as tokens and establishes full correspondence (as shown in Fig. \ref{fig:meth_vis} (b)), it is called all-to-all attention. Although all-to-all attention has been a key factor in many methods \cite{park2019arbitrary, yao2019attention,liu2021adaattn}, its defects have not yet been found and studied. We argue that the all-to-all attention form is inadequate for AST due to its inherent limitations, as analyzed in Introduction \ref{sec:introduction}.

\textbf{Distributed Attention.} To alleviate the problem caused by all-to-all attention, distributed attention first learns \emph{distributed} keys that depict the style distribution of all local regions of the style features. Then, each query of the content features is matched with these representative keys (as shown in Fig. \ref{fig:meth_vis} (e)). Distributed attention can improve matching error tolerance since the matched keys are regional style representations rather than isolated positions. Even if the exclusivity of $softmax$ causes the attention to only focus on the most similar key, the most similar key can also reflect the style information of a region so that it will not directly match the isolated style pattern (as shown in Fig. \ref{fig:meth_vis} (f)). In addition, since these keys are fixed to depict several local regions, the matching effect of DA has high stability for queries in different positions. Technically, we implement distributed attention in a blocked, sparse fashion, as shown in Fig. \ref{fig:meth_dp}. The content feature $F_c^l$ and the style feature $F_s^l$ with the size of $(H^l, W^l, C^l)$ at each layer $l$ are spatially blocked into tensors that stand for $Q^l$ and $K^l$ respectively:
\begin{equation}
Q^l = B^l(N^l(F_c^l)), K^l = B^l(N^l(F_s^l)),
\end{equation}
where $N^l(*)$ denotes instance normalization and $B^l(*)$ denotes $1 \times 1$ Conv-Blocking operation. The shape of $Q^l$ and $K^l$ is $(b^l \times b^l, \frac{H^l}{b^l} \times \frac{W^l}{b^l}, C^l)$, representing $b^l \times b^l$ non-overlapping blocks each with the size of $(\frac{H^l}{b^l}, \frac{W^l}{b^l})$. Therefore, each block contains $n = \frac{H^l}{b^l} \times \frac{W^l}{b^l}$ points. We denote the points in each block as $k_i$. The first step of distributed attention is regional style aggregation. We calculate the mean of all points in each block as the initial style representation:
\begin{equation}
k_m = \frac{1}{n}\sum_{i=1}^n k_i.
\end{equation}
Then, we dynamically aggregate all points in a block based on the similarities to the mean point. Assuming the similarity between the $n$ points and the mean point is $s \in \mathfrak{R}^n$, the aggregated regional style representation is given by:
\begin{equation}
k = \frac{1}{n}(k_m + \sum_{i=1}^n \rm{sig}(\alpha s_i + \beta) k_i),
\end{equation}
where $\alpha$ and $\beta$ are learnable scalars to scale and shift the similarity and $\rm{sig}$ is a sigmoid function to re-scale the similarity to $(0, 1)$. After aggregation, we obtain $b^l \times b^l$ new keys, denoted as $\widehat{K}^l$. Note that we also adopt the same aggregation method for $B(F_s^l)$ to get new values, expressed as $\widehat{B}(F_s^l)$.
In the second step, distributed attention $D^l$ calculates point-wise attention on the first axis of $Q^l$ and $\widehat{K}^l$:
\begin{equation}
D_{attn}^l = D^l(Q^l, \widehat{K}^l).
\end{equation}
The attention score matrix $D_{attn}^l$, with the size of $(\frac{H^l}{b^l} \times \frac{W^l}{b^l}, b^l \times b^l, b^l \times b^l)$, stores the similarity correspondence of point-wise tokens across $b^l \times b^l$ blocks and indexed in range $\frac{H^l}{b^l} \times \frac{W^l}{b^l}$. Note that attention along a specific axis of $Q^l$ and $K^l$ can be realized straightforwardly by einsum operation, which most deep learning frameworks have implemented. In addition, inspired by the multi-head attention in transformer \cite{vaswani2017attention}, we use multiple heads to split the tensors along channel dimensions and project the divided tensors into different spaces for attention calculation.

\textbf{Progressive Attention.} All-to-all attention focuses directly on a specific position. PA \emph{progressively} concentrates its attention from the coarse-grained region to the fine-grained position (as shown in Fig. \ref{fig:meth_vis} (g) (h) (i)). Paying attention to the coarse-grained (as shown in Fig. \ref{fig:meth_vis} (g)) in the first step contributes to matching style patterns on a larger scale; thus, the attention after $argmax$ (as shown in Fig. \ref{fig:meth_vis} (h)) can concentrate on a coarse style pattern with more similar semantics. Fine-grained positions can be further located via point-wise attention within this coarse-grained region, as shown in Fig. \ref{fig:meth_vis}. In addition, since queries within a local region are matched to the same keys within a local region, their transferred features also have regional stability. The technical implementation of PA is shown in Fig. \ref{fig:meth_dp}. $Q^l$ and $K^l$ are obtained similarly as distributed attention but using a different $1 \times 1$ Conv. The first step of progressive attention is implemented as patch-wise attention along the first axis, which takes a block region as a token instead of a specific position. Specifically, we use $argmax$ to match only the most similar coarse-grained region, and therefore, the output of this step is the index matrix that stores sparse indices of patch-wise tokens across $b^l \times b^l$ blocks:
\begin{equation}
P_{idx}^l = argmax(P_{1}^l(Q^l, K^l)),
\end{equation} 
where $P_{1}^l$ denotes the first step of PA. With $P_{idx}^l$, we can reshuffle the tokens of $K^l$ and $B(F_s^l)$ to semantically matching the spatial arrangement of the tokens of $Q^l$:
\begin{equation}
\begin{aligned}
\widetilde{K}^l &= reshuffle(K^l, P_{idx}^l), \\
\widetilde{B}(F_s^l) &= reshuffle(B(F_s^l), P_{idx}^l),
\end{aligned}
\end{equation} 
where $reshuffle(*,*)$ denotes the reshuffle operation. The second step of progressive attention $P_{2}^l$ is implemented as regional attention, where tokens attend to their neighbors within non-overlapped blocks. Attention score in this step is calculated along the second axis of $Q^l$ and $\widetilde{K}^l$:
\begin{equation}
P_{attn}^l = P_{2}^l(Q^l, \widetilde{K}^l).
\end{equation} 
The attention score matrix $P_{attn}^l$, with the size of $(b^l \times b^l, \frac{H^l}{b^l} \times \frac{W^l}{b^l}, \frac{H^l}{b^l} \times \frac{W^l}{b^l})$, stores the sparse similarity correspondence of point-wise tokens within blocks with the size of $\frac{H^l}{b^l} \times \frac{W^l}{b^l}$ indexed in the range $b^l \times b^l$. Both steps of progressive attention can be implemented with einsum.

\textbf{Feature Transformation.} With the output attention score, the feature transformation can be performed by:
\begin{equation}
\begin{aligned}
&F_{cs\_D}^l = U(D_{attn}^l \cdot \widehat{B}(F_s^l)), \\
&F_{cs\_P}^l = U(P_{attn}^l \cdot \widetilde{B}(F_s^l)),
\end{aligned}
\end{equation}
where $U(*)$ denotes the Unblocking-Conv operation, $\cdot$ represents the dot product between a specific axis of two tensors which can be implemented with einsum. The transferred feature at each layer $l$ can be eventually obtained by: 
\begin{equation}
F_{cs}^l = F_{cs\_D}^l + F_{cs\_P}^l + F_c^l,
\end{equation}
where $F_{cs\_D}^l$, $F_{cs\_P}^l$, and $F_c^l$ are the output of the distributed attention path, the output of the progressive attention path, and the content feature, respectively. 

\textbf{Complexity Analysis.} The computational complexity of A2K is:
\begin{equation}
\begin{aligned}
\Omega &= \Omega_{D_1} + \Omega_{D_2} + \Omega_{P_1} + \Omega_{P_{2}} \\
&= [5 + (b)^2 + (b)^2 + \frac{H}{b}\frac{W}{b}] HWC, 
\end{aligned}
\end{equation}
which saves $\mathcal{O}(\sqrt{HW})$ computational consumption with respect to image size HW when $b^2 \approx \frac{H}{b}\frac{W}{b}$.

\subsection{Loss Function}

The loss function for training the model consists of two terms. One of them is the style loss $\mathcal{L}_{gs}$ followed by \cite{huang2017arbitrary}, which penalizes the Euclidean distances of mean $\mu $ and standard deviation $\sigma$ between stylized image and style image in VGG feature space to ensure global stylization effect:
\begin{equation}
\begin{aligned}
\mathcal{L}_{gs}
& = \sum_{l=1}^{4}\vert\vert\mu(E_{nc}^l(I_{cs}))-\mu(F_s^l)\vert\vert_2 \\
& + \sum_{l=1}^{4}\vert\vert\sigma(E_{nc}^l(I_{cs}))-\sigma(F_s^l)\vert\vert_2,
\end{aligned}
\end{equation}
where $E_{nc}^l(*)$ denotes feature extracted from the $l$th layer of the pre-trained VGG encoder. The second term is an attention-based feature matching loss that penalizes the Euclidean distance between the transferred features of the A2K module and the features of the stylized image:
\begin{equation}
\mathcal{L}_{A2K^*} = \sum_{l=2}^{5}\vert\vert E_{nc}^l(I_{cs})-M_{A2K^*}^l(F_c^l,F_s^l) \vert\vert_2,
\end{equation}
where $M_{A2K^*}^l$ denotes a non-parametric version that removes the learnable $1 \times 1$ Conv because the supervision signal should be deterministic. The full loss is:
\begin{equation}
\label{eq:loss}
\mathcal{L} = \lambda_1 \mathcal{L}_{gs} + \lambda_2 \mathcal{L}_{A2K^*}.
\end{equation}
We empirically set $\lambda_1$ and $\lambda_2$ as 10 and 1.25. See supplementary material for a detailed analysis.


\section{Experiments}
\label{sec:experiments}

\subsection{Implementing Details}

We use images from MS-COCO \cite{lin2014microsoft} as content and images from WikiArt \cite{phillips2011wiki} as style to train our model. The head number is set to 8. The batch size is set to 8, and the training lasts for five epochs (400K iterations) on a single NVIDIA GeForce RTX 3090 GPU. Adam \cite{kingma2014adam} with momentum parameters $\beta_1 = 0.9$ and $\beta_2 = 0.999$ is used for optimization. The learning rate is set to 2e-4 for the first two epochs and linearly decayed to 0 for the rest three epochs. Images are randomly cropped to 256 $\times$ 256 during training and loaded with 512 $\times$ 512 during inference. To ensure $b^l \times b^l \approx \frac{H^l}{b^l} \times \frac{W^l}{b^l}$, the block size $b$ is set to 16, 8, 8, and 4 for $ReLU 2\_1$, $3\_1$, $4\_1$ and $5\_1$ layers, respectively. 

\subsection{Comparison with Prior Arts}

To validate the effectiveness of the proposed method, we compare it with six previous state-of-the-art AST methods: $\mathrm{StyTr^2}$ \cite{deng2022stytr2}, AdaAttN \cite{liu2021adaattn}, SANet \cite{park2019arbitrary}, MST \cite{zhang2019multimodal}, Avatar-Net \cite{sheng2018avatar}, and AdaIN \cite{huang2017arbitrary}. Among them, SANet and AdaAttN are typically all-to-all attention-based methods. We obtain the results of the comparison methods by running their official code with their default configurations.

\begin{figure*}[t]
\centering
\includegraphics[width=0.90\linewidth]{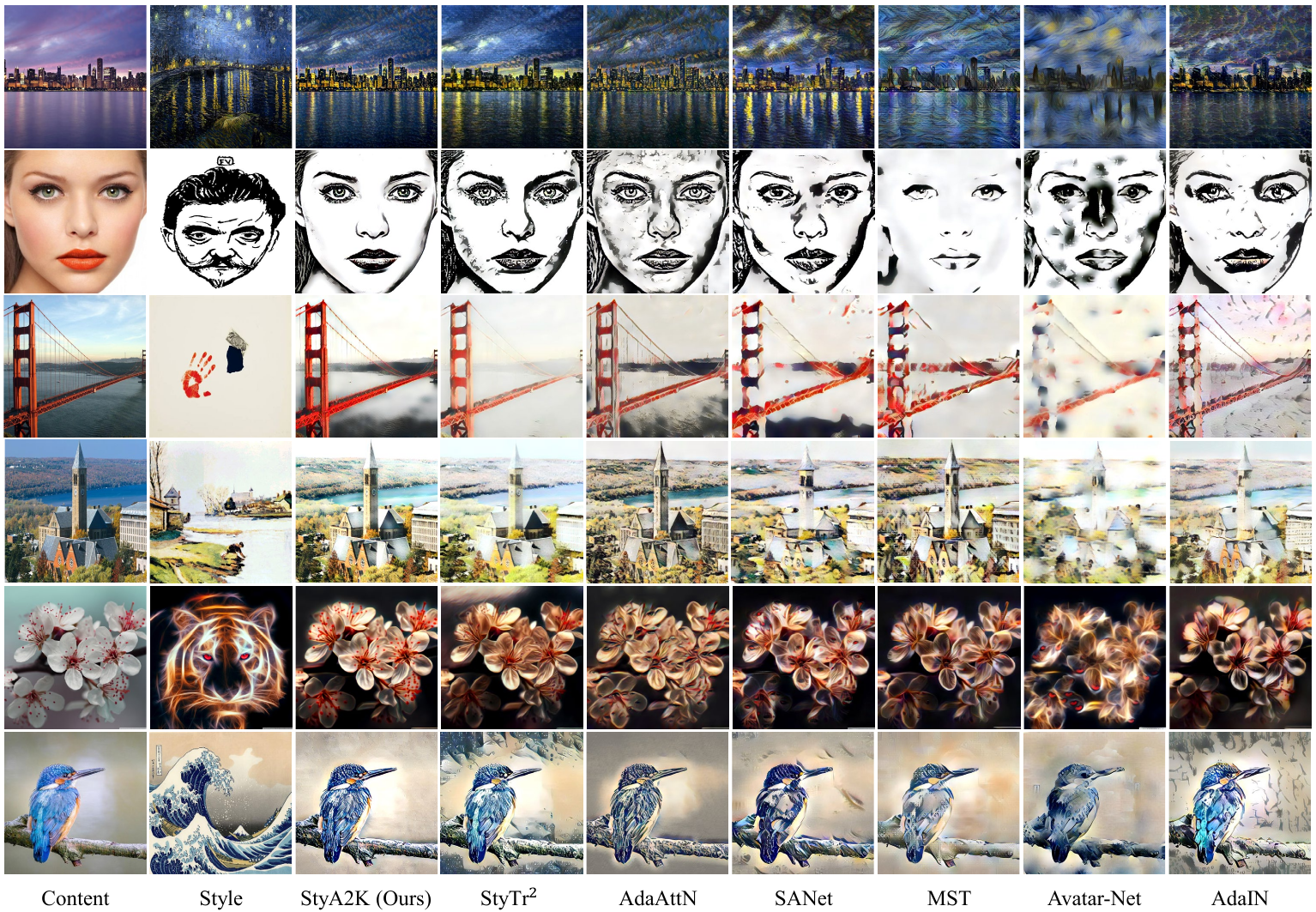}
\caption{Visual comparisons among different AST methods. Zoom in for a better view.}
\label{fig:expr_sota}
\end{figure*}

\begin{table*}[t]
\centering
\caption{Statics of quantitative comparison, user study, and inference time. The best results are in \textbf{bold} and the second best results are marked with an \underline{underline}.}
\scalebox{0.90}{
\begin{tabular}{c|c|ccccccc} 
\toprule
\multicolumn{2}{c|}{Method} & StyA2K (Ours)    & $\mathrm{StyTr^2}$ & AdaAttN           & SANet              & MST              & Avatar-Net & AdaIN  \\
\hline 
\multirow{3}{*}{Evaluation Metrics} 
& Content Loss $\downarrow$ & \textbf{0.55}    & 0.83               & 1.00              & 0.96               & \underline{0.77} & 1.26       & 0.91   \\
& Style Loss $\downarrow$   & \underline{1.04}    & 1.20               & 1.24              & \textbf{0.99}   & 2.65             & 3.96       & 1.16   \\
& LPIPS $\downarrow$        & \textbf{0.52}    & 0.57               & 0.59              & 0.63               & \underline{0.57} & 0.64       & 0.62   \\
\hline 
\multirow{3}{*}{User Study}
& Content Per. $\uparrow$   & \textbf{52.69}   & 9.81               & \underline{15.56} & 7.12               & 8.19             & 2.25       & 4.38   \\
& Style Per. $\uparrow$     & \textbf{29.06}   & 14.81              & 12.38             & \underline{16.13}  & 11.31            & 9.13       & 7.19   \\
& Overall Per. $\uparrow$   & \textbf{36.06}   & \underline{15.81}  & 14.50             & 11.44              & 11.63            & 3.56       & 7.00   \\
\hline 
\multicolumn{2}{c|}{Inference Time (Sec./Image) $\downarrow$}
                              & 0.0143           & 0.1004             & 0.0269            & \underline{0.0041} & 1.3906           & 0.3348     & \textbf{0.0038} \\
\bottomrule
\end{tabular}}
\label{tab:expr_quan}
\vspace{-4mm}
\end{table*}

\textbf{Qualitative Comparisons.} We show the visual comparisons between our model and other AST methods in Fig. \ref{fig:meth_vis}. AdaIN adjusts the holistic feature distribution and has no local content awareness; therefore, its results can only roughly transfer the holistic style. Avatar-Net utilizes a relaxed normalized cross-correlation to fulfill locality-aware feature matching so that its results partially maintain semantic consistency. The results are significantly distorted, though, because Avatar-Net only matches the most similar style pattern. MST clusters the complex style distribution into sub-style components and manipulates the features by graph-based patch matching. Despite its promising results, distorted structures still exist. Additionally, it heavily relies on the effect of clustering. SANet adopts an attention mechanism to match style features for local content features attentively. AdaAttN combines attention mechanism and adaptive instance normalization to integrate the advantage of locality-aware and holistic feature matching. Their results show fine local style details. However, due to the limitation of all-to-all attention, their results contain many unreasonable, inconsistent style patterns (2nd, 3rd, and 5th row). $\mathrm{StyTr^2}$ adapts transformer architecture to AST and achieves cutting-edge performance. Their results still have unreasonable textures (2nd, 4td, and 6th row). Our proposed StyA2K outperforms other methods in maintaining semantic structures and generating consistent style patterns.

\textbf{Quantitative Comparisons.} Following \cite{deng2022stytr2}, we compute the average content loss between the generated results and input content images and the average style loss between the generated results and input style images to measure how well the input content and style are preserved. The smaller the value is, the better the input content/style is preserved. In addition, we adopt the Learned Perceptual Image Patch Similarity (LPIPS) \cite{zhang2018unreasonable} metric to calculate the quality difference between the generated and input content images. A lower score indirectly indicates better quality of generated images. We randomly selected 15 content and 20 style images to generate 300 stylized images. Table \ref{tab:expr_quan} shows the corresponding quantitative results. StyA2K shows superior performance on content loss and LPIPS score, which indicates its superiority in preserving semantic structures and rendering consistent style patterns. The style loss of StyA2K's results is slightly higher than SANet but lower than other methods. Therefore, our results can significantly improve the structure and texture consistency while achieving comparable style performance.

\textbf{User Study.} We conduct a user study to evaluate human preference on the results generated by different methods. We reuse the images in the quantitative comparisons and randomly sample 20 groups. Each group consists of a content image, a style image, and seven stylized images generated by different methods. The order of stylized images in each group is shuffled. We ask participants to select their favorite stylized image in each group from three perspectives: content preservation, stylization effect, and overall preference. We collect 1600 votes for each view from 80 participants. Table \ref{tab:expr_quan} shows the statistics of the votes, which indicate that the results generated by our method are more appealing than other methods.

\textbf{Running time Comparison.} We report the average inference time of StyA2K and other AST methods under $512 \times 512$ resolution in Table \ref{tab:expr_quan}. StyA2K achieves 70 FPS at $512 \times 512$ resolution, which can run in real time. StyA2K is faster than AdaAttN. The reason why StyA2K is slower than SANet is that SANet only performs feature transformation on two layers of VGG features.

\subsection{Ablation Study}

\begin{figure}[t]
\centering
\includegraphics[width=0.96\linewidth]{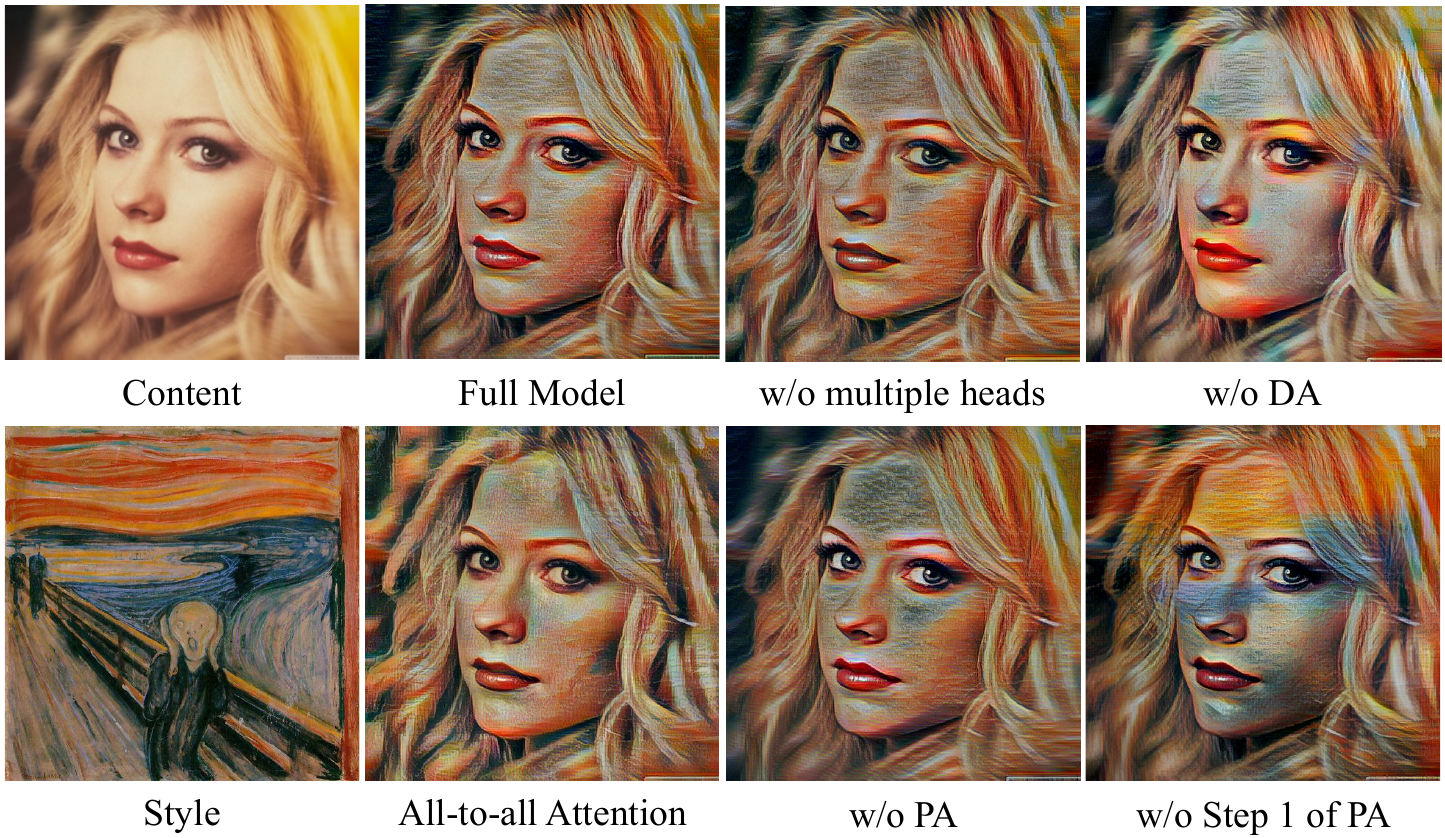}
\caption{Ablation study on the key ingredients of StyA2K.}
\label{fig:expr_abla}
\end{figure}

We conduct an ablation study to verify the effectiveness of the key ingredients in our method. Visual and quantitative results are shown in Fig. \ref{fig:expr_abla} and Table \ref{tab:expr_abla}, respectively.

\textbf{Effect of All-to-key Attention.} To demonstrate the superiority of our proposed all-to-key attention over all-to-all attention, we replace the all-to-key attention in our full model with all-to-all attention and observe the changes in visual quality and evaluation scores. As shown in Fig. \ref{fig:expr_abla}, the model using all-to-all attention produces distorted and inconsistent style patterns in its result. The evaluation scores of all-to-all attention's results are all worse than all-to-key attention, as shown in Table \ref{tab:expr_abla}. The results suggest that all-to-key attention alleviates the defects caused by all-to-all attention and thus achieves superior performance.

\textbf{Effect of DA and PA.} To validate the respective effectiveness of DA and PA, we set up three variants: 1) full model without DA, 2) full model without PA, and 3) full model without the first step of PA. The results of these variants are shown in \ref{fig:expr_abla}. The model without DA can match correct style patterns with high semantic similarity but lack in rendering spatial consistent style patterns. The model without PA has high style consistency in local regions but produces patterns with incorrect semantics. When removing the first step, PA directly performs regional attention within the same location. In other words, it directly introduces the content semantics from the reference style. Therefore, its result is mixed with the spatially copied style patterns, as shown in \ref{fig:expr_abla}, and its results obtain the lowest style loss, as shown in Table \ref{tab:expr_abla}. However, introducing content semantics from the reference style image deviates from the setting of the style transfer task. In addition, almost all the average content loss and the average LPIPS score increase when removing one of the three components. Therefore, we can conclude that all three ingredients are critical to the final effect of the model.

\textbf{Effect of Multiple Heads.} The removal of multiple heads (i.e., using only one head) leads to slight visual quality degradation and loss increase, as shown in Fig. \ref{fig:expr_abla} and Table \ref{tab:expr_abla}. We can conclude that multiple heads help to improve the final performance of the proposed method.

\begin{table}[t]
\centering
\caption{Ablation study on the key ingredients of the proposed method. The best results are in \textbf{bold} and the second best results are marked with an \underline{underline}.} 
\scalebox{0.82}{
\begin{tabular}{c|ccc} 
\toprule
Model                        & Content Loss $\downarrow$ & Style Loss $\downarrow$ & LPIPS $\downarrow$ \\
\hline 
All-to-all Attention         & 0.73                      & 1.16                    & 0.56 \\
\hline 
w/o DA                       & 0.59                      & 1.48                    & \textbf{0.51} \\
w/o PA                       & \underline{0.58}          & \underline{1.00}        & 0.59 \\
w/o Step 1 of PA             & 0.65                      & \textbf{0.73}           & 0.56 \\
\hline
w/o multiple heads           & 0.65                      & 1.05                    & 0.54 \\
\hline
Full Model                   & \textbf{0.55}             & 1.04                    & \underline{0.52} \\
\bottomrule
\end{tabular}}
\label{tab:expr_abla}
\vspace{-4mm}
\end{table}

\subsection{Multi-style Transfer}

\begin{figure}[t]
\centering
\includegraphics[width=0.86\linewidth]{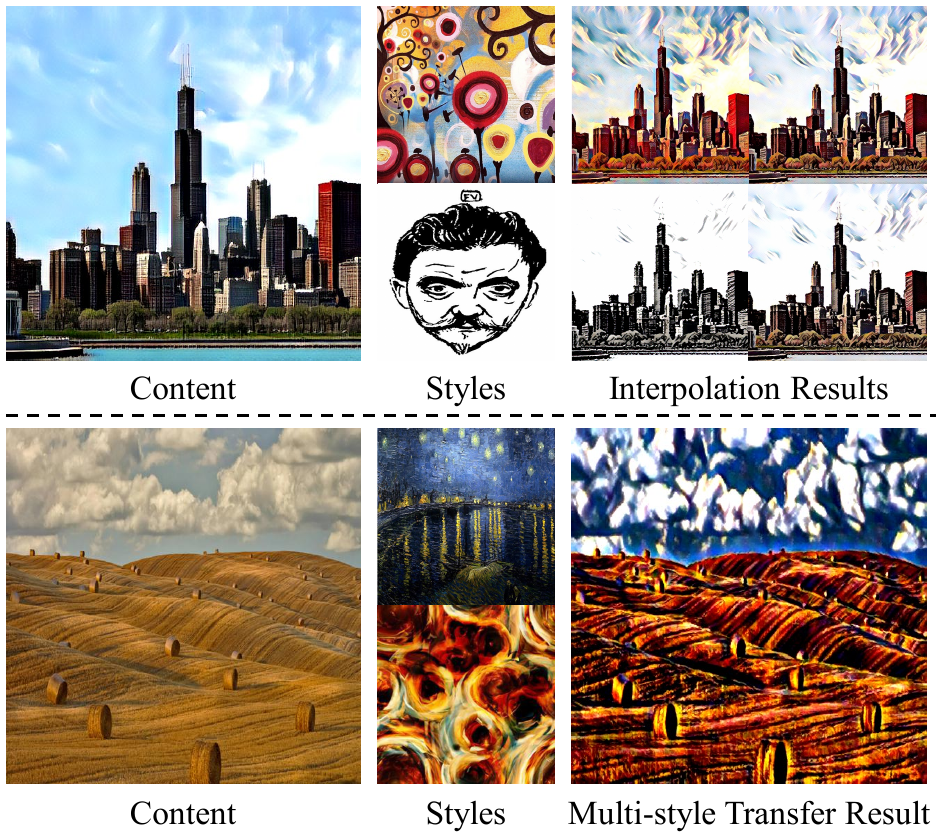}
\caption{Results of style interpolation and multi-style transfer.}
\label{fig:expr_multi}
\end{figure}

Following previous studies \cite{deng2020arbitrary, park2019arbitrary,liu2021adaattn}, StyA2K can achieve style interpolation between different styles and accomplish multi-style transfer that integrates multiple styles in one output image, as shown in Fig. \ref{fig:expr_multi}. These results demonstrate the flexibility of StyA2K.

\subsection{Video Style Transfer}

Since A2K has excellent stability to position changes (such as object motion and light changes), we can directly apply it to video style transfer. We do not add any mechanism based on the video prior, do not conduct any training on video data, and independently process each frame. Fig. \ref{fig:intro_advan} shows a stylized video by our method. See supplementary material for more results. We also directly apply AdaAttN ($softmax$ version) to video style transfer. Through comparison, we can find that our method has almost no flicker phenomenon and shows high temporal consistency. We calculate the optical flow error \cite{chen2017coherent} on five stylized video clips to quantitatively verify the stability of our method. As shown in Tab. \ref{tab:expr_video}, StyA2K has a lower optical flow error than AdaAttN in all five stylized video clips. Note that we have no intention to propose a SOTA video style transfer method because it requires introducing other technologies, such as temporal consistency constraint. We directly apply StyA2K to video style transfer to verify its stability.

\begin{table}[t]
\centering
\caption{Optical flow error of AdaAttN and StyA2K. Lower values indicate better temporal consistency.} 
\scalebox{0.82}{
\begin{tabular}{c|cccccc} 
\toprule
Model           & Style1  & Style2  & Style3 & Style4 & Style5 & Mean  \\
\hline 
AdaAttN         & 3.03    & 3.46    & 6.00   &  3.78  & 8.45   & 4.94  \\
\hline 
StyA2K          & 2.58    & 2.56    & 4.99   &  3.21  & 7.23   & 4.24  \\
\bottomrule
\end{tabular}}
\label{tab:expr_video}
\vspace{-4mm}
\end{table}

\section{Conclusion and Limitation}

This paper proposes a novel all-to-key attention mechanism to achieve effective and efficient arbitrary style transfer. It integrates two inventions: distributed attention and progressive attention. Distributed attention assigns attention to key style representations. Progressive attention pays attention from coarse to fine. They jointly promote maintaining semantic structures and synthesizing spatially consistent style texture. Extensive experiments verify the superiority of our method.

\textbf{Limitation.} In this study, we are committed to solving the problems caused by all-to-all attention and improving the semantic structure preservation, style pattern consistency, and stability of style transfer. However, we have not explored the aspect of improving style expressiveness. Recently, contrastive learning has shown strong performance in learning style representation \cite{zhang2022domain}. Therefore, we will explore the feasibility of combining all-to-key attention with contrastive learning in our future work to improve the style expression of the method.

{\small
\bibliographystyle{ieee_fullname}
\bibliography{stya2k}

\begin{thebibliography}{10}\itemsep=-1pt

\bibitem{an2021artflow}
Jie An, Siyu Huang, Yibing Song, Dejing Dou, Wei Liu, and Jiebo Luo.
\newblock Artflow: Unbiased image style transfer via reversible neural flows.
\newblock In {\em Proceedings of the IEEE/CVF Conference on Computer Vision and
  Pattern Recognition}, pages 862--871, 2021.

\bibitem{chen2017coherent}
Dongdong Chen, Jing Liao, Lu Yuan, Nenghai Yu, and Gang Hua.
\newblock Coherent online video style transfer.
\newblock In {\em Proceedings of the IEEE International Conference on Computer
  Vision}, pages 1105--1114, 2017.

\bibitem{chen2017stylebank}
Dongdong Chen, Lu Yuan, Jing Liao, Nenghai Yu, and Gang Hua.
\newblock Stylebank: An explicit representation for neural image style
  transfer.
\newblock In {\em Proceedings of the IEEE conference on computer vision and
  pattern recognition}, pages 1897--1906, 2017.

\bibitem{chen2016fast}
Tian~Qi Chen and Mark Schmidt.
\newblock Fast patch-based style transfer of arbitrary style.
\newblock {\em arXiv preprint arXiv:1612.04337}, 2016.

\bibitem{cheng2021style}
Jiaxin Cheng, Ayush Jaiswal, Yue Wu, Pradeep Natarajan, and Prem Natarajan.
\newblock Style-aware normalized loss for improving arbitrary style transfer.
\newblock In {\em Proceedings of the IEEE/CVF Conference on Computer Vision and
  Pattern Recognition}, pages 134--143, 2021.

\bibitem{deng2022stytr2}
Yingying Deng, Fan Tang, Weiming Dong, Chongyang Ma, Xingjia Pan, Lei Wang, and
  Changsheng Xu.
\newblock Stytr2: Image style transfer with transformers.
\newblock In {\em Proceedings of the IEEE/CVF Conference on Computer Vision and
  Pattern Recognition}, pages 11326--11336, 2022.

\bibitem{deng2020arbitrary}
Yingying Deng, Fan Tang, Weiming Dong, Wen Sun, Feiyue Huang, and Changsheng
  Xu.
\newblock Arbitrary style transfer via multi-adaptation network.
\newblock In {\em Proceedings of the 28th ACM international conference on
  multimedia}, pages 2719--2727, 2020.

\bibitem{dosovitskiy2020image}
Alexey Dosovitskiy, Lucas Beyer, Alexander Kolesnikov, Dirk Weissenborn,
  Xiaohua Zhai, Thomas Unterthiner, Mostafa Dehghani, Matthias Minderer, Georg
  Heigold, Sylvain Gelly, et~al.
\newblock An image is worth 16x16 words: Transformers for image recognition at
  scale.
\newblock {\em arXiv preprint arXiv:2010.11929}, 2020.

\bibitem{dumoulin2016learned}
Vincent Dumoulin, Jonathon Shlens, and Manjunath Kudlur.
\newblock A learned representation for artistic style.
\newblock {\em arXiv preprint arXiv:1610.07629}, 2016.

\bibitem{gatys2016image}
Leon~A Gatys, Alexander~S Ecker, and Matthias Bethge.
\newblock Image style transfer using convolutional neural networks.
\newblock In {\em Proceedings of the IEEE conference on computer vision and
  pattern recognition}, pages 2414--2423, 2016.

\bibitem{gatys2017controlling}
Leon~A Gatys, Alexander~S Ecker, Matthias Bethge, Aaron Hertzmann, and Eli
  Shechtman.
\newblock Controlling perceptual factors in neural style transfer.
\newblock In {\em Proceedings of the IEEE conference on computer vision and
  pattern recognition}, pages 3985--3993, 2017.

\bibitem{gu2018arbitrary}
Shuyang Gu, Congliang Chen, Jing Liao, and Lu Yuan.
\newblock Arbitrary style transfer with deep feature reshuffle.
\newblock In {\em Proceedings of the IEEE Conference on Computer Vision and
  Pattern Recognition}, pages 8222--8231, 2018.

\bibitem{huang2017arbitrary}
Xun Huang and Serge Belongie.
\newblock Arbitrary style transfer in real-time with adaptive instance
  normalization.
\newblock In {\em Proceedings of the IEEE international conference on computer
  vision}, pages 1501--1510, 2017.

\bibitem{jiang2020psgan}
Wentao Jiang, Si Liu, Chen Gao, Jie Cao, Ran He, Jiashi Feng, and Shuicheng
  Yan.
\newblock Psgan: Pose and expression robust spatial-aware gan for customizable
  makeup transfer.
\newblock In {\em Proceedings of the IEEE/CVF Conference on Computer Vision and
  Pattern Recognition}, pages 5194--5202, 2020.

\bibitem{jing2020dynamic}
Yongcheng Jing, Xiao Liu, Yukang Ding, Xinchao Wang, Errui Ding, Mingli Song,
  and Shilei Wen.
\newblock Dynamic instance normalization for arbitrary style transfer.
\newblock In {\em Proceedings of the AAAI Conference on Artificial
  Intelligence}, pages 4369--4376, 2020.

\bibitem{johnson2016perceptual}
Justin Johnson, Alexandre Alahi, and Li Fei-Fei.
\newblock Perceptual losses for real-time style transfer and super-resolution.
\newblock In {\em European conference on computer vision}, pages 694--711.
  Springer, 2016.

\bibitem{kalischek2021light}
Nikolai Kalischek, Jan~D Wegner, and Konrad Schindler.
\newblock In the light of feature distributions: moment matching for neural
  style transfer.
\newblock In {\em Proceedings of the IEEE/CVF Conference on Computer Vision and
  Pattern Recognition}, pages 9382--9391, 2021.

\bibitem{kingma2014adam}
Diederik~P Kingma and Jimmy Ba.
\newblock Adam: A method for stochastic optimization.
\newblock {\em arXiv preprint arXiv:1412.6980}, 2014.

\bibitem{kotovenko2019disentanglement}
Dmytro Kotovenko, Artsiom Sanakoyeu, Sabine Lang, and Bjorn Ommer.
\newblock Content and style disentanglement for artistic style transfer.
\newblock In {\em Proceedings of the IEEE/CVF international conference on
  computer vision}, pages 4422--4431, 2019.

\bibitem{li2016combining}
Chuan Li and Michael Wand.
\newblock Combining markov random fields and convolutional neural networks for
  image synthesis.
\newblock In {\em Proceedings of the IEEE conference on computer vision and
  pattern recognition}, pages 2479--2486, 2016.

\bibitem{li2016precomputed}
Chuan Li and Michael Wand.
\newblock Precomputed real-time texture synthesis with markovian generative
  adversarial networks.
\newblock In {\em European conference on computer vision}, pages 702--716.
  Springer, 2016.

\bibitem{li2019learning}
Xueting Li, Sifei Liu, Jan Kautz, and Ming-Hsuan Yang.
\newblock Learning linear transformations for fast image and video style
  transfer.
\newblock In {\em Proceedings of the IEEE/CVF Conference on Computer Vision and
  Pattern Recognition}, pages 3809--3817, 2019.

\bibitem{li2017diversified}
Yijun Li, Chen Fang, Jimei Yang, Zhaowen Wang, Xin Lu, and Ming-Hsuan Yang.
\newblock Diversified texture synthesis with feed-forward networks.
\newblock In {\em Proceedings of the IEEE conference on computer vision and
  pattern recognition}, pages 3920--3928, 2017.

\bibitem{li2017universal}
Yijun Li, Chen Fang, Jimei Yang, Zhaowen Wang, Xin Lu, and Ming-Hsuan Yang.
\newblock Universal style transfer via feature transforms.
\newblock {\em Advances in neural information processing systems}, 30, 2017.

\bibitem{liao2017visual}
Jing Liao, Yuan Yao, Lu Yuan, Gang Hua, and Sing~Bing Kang.
\newblock Visual attribute transfer through deep image analogy.
\newblock {\em arXiv preprint arXiv:1705.01088}, 2017.

\bibitem{lin2014microsoft}
Tsung-Yi Lin, Michael Maire, Serge Belongie, James Hays, Pietro Perona, Deva
  Ramanan, Piotr Doll{\'a}r, and C~Lawrence Zitnick.
\newblock Microsoft coco: Common objects in context.
\newblock In {\em Proceedings of the European Conference on Computer Vision
  (ECCV)}, pages 740--755. Springer, 2014.

\bibitem{liu2021adaattn}
Songhua Liu, Tianwei Lin, Dongliang He, Fu Li, Meiling Wang, Xin Li, Zhengxing
  Sun, Qian Li, and Errui Ding.
\newblock Adaattn: Revisit attention mechanism in arbitrary neural style
  transfer.
\newblock In {\em Proceedings of the IEEE/CVF international conference on
  computer vision}, pages 6649--6658, 2021.

\bibitem{liu2021swin}
Ze Liu, Yutong Lin, Yue Cao, Han Hu, Yixuan Wei, Zheng Zhang, Stephen Lin, and
  Baining Guo.
\newblock Swin transformer: Hierarchical vision transformer using shifted
  windows.
\newblock In {\em Proceedings of the IEEE/CVF International Conference on
  Computer Vision}, pages 10012--10022, 2021.

\bibitem{park2019arbitrary}
Dae~Young Park and Kwang~Hee Lee.
\newblock Arbitrary style transfer with style-attentional networks.
\newblock In {\em proceedings of the IEEE/CVF conference on computer vision and
  pattern recognition}, pages 5880--5888, 2019.

\bibitem{phillips2011wiki}
Fred Phillips and Brandy Mackintosh.
\newblock Wiki art gallery, inc.: A case for critical thinking.
\newblock {\em Issues in Accounting Education}, 26(3):593--608, 2011.

\bibitem{shen2018neural}
Falong Shen, Shuicheng Yan, and Gang Zeng.
\newblock Neural style transfer via meta networks.
\newblock In {\em Proceedings of the IEEE Conference on Computer Vision and
  Pattern Recognition}, pages 8061--8069, 2018.

\bibitem{sheng2018avatar}
Lu Sheng, Ziyi Lin, Jing Shao, and Xiaogang Wang.
\newblock Avatar-net: Multi-scale zero-shot style transfer by feature
  decoration.
\newblock In {\em Proceedings of the IEEE conference on computer vision and
  pattern recognition}, pages 8242--8250, 2018.

\bibitem{simonyan2014very}
Karen Simonyan and Andrew Zisserman.
\newblock Very deep convolutional networks for large-scale image recognition.
\newblock {\em arXiv preprint arXiv:1409.1556}, 2014.

\bibitem{ulyanov2016texture}
Dmitry Ulyanov, Vadim Lebedev, Andrea Vedaldi, and Victor~S Lempitsky.
\newblock Texture networks: Feed-forward synthesis of textures and stylized
  images.
\newblock In {\em International Conference on Machine Learning}, page~4, 2016.

\bibitem{ulyanov2017improved}
Dmitry Ulyanov, Andrea Vedaldi, and Victor Lempitsky.
\newblock Improved texture networks: Maximizing quality and diversity in
  feed-forward stylization and texture synthesis.
\newblock In {\em Proceedings of the IEEE conference on computer vision and
  pattern recognition}, pages 6924--6932, 2017.

\bibitem{vaswani2021scaling}
Ashish Vaswani, Prajit Ramachandran, Aravind Srinivas, Niki Parmar, Blake
  Hechtman, and Jonathon Shlens.
\newblock Scaling local self-attention for parameter efficient visual
  backbones.
\newblock In {\em Proceedings of the IEEE/CVF Conference on Computer Vision and
  Pattern Recognition}, pages 12894--12904, 2021.

\bibitem{vaswani2017attention}
Ashish Vaswani, Noam Shazeer, Niki Parmar, Jakob Uszkoreit, Llion Jones,
  Aidan~N Gomez, {\L}ukasz Kaiser, and Illia Polosukhin.
\newblock Attention is all you need.
\newblock {\em Advances in neural information processing systems}, 30, 2017.

\bibitem{wang2018non}
Xiaolong Wang, Ross Girshick, Abhinav Gupta, and Kaiming He.
\newblock Non-local neural networks.
\newblock In {\em Proceedings of the IEEE conference on computer vision and
  pattern recognition}, pages 7794--7803, 2018.

\bibitem{wang2017multimodal}
Xin Wang, Geoffrey Oxholm, Da Zhang, and Yuan-Fang Wang.
\newblock Multimodal transfer: A hierarchical deep convolutional neural network
  for fast artistic style transfer.
\newblock In {\em Proceedings of the IEEE conference on computer vision and
  pattern recognition}, pages 5239--5247, 2017.

\bibitem{wu2021styleformer}
Xiaolei Wu, Zhihao Hu, Lu Sheng, and Dong Xu.
\newblock Styleformer: Real-time arbitrary style transfer via parametric style
  composition.
\newblock In {\em Proceedings of the IEEE/CVF International Conference on
  Computer Vision}, pages 14618--14627, 2021.

\bibitem{yang2019xlnet}
Zhilin Yang, Zihang Dai, Yiming Yang, Jaime Carbonell, Russ~R Salakhutdinov,
  and Quoc~V Le.
\newblock Xlnet: Generalized autoregressive pretraining for language
  understanding.
\newblock {\em Advances in neural information processing systems}, 32, 2019.

\bibitem{yao2019attention}
Yuan Yao, Jianqiang Ren, Xuansong Xie, Weidong Liu, Yong-Jin Liu, and Jun Wang.
\newblock Attention-aware multi-stroke style transfer.
\newblock In {\em Proceedings of the IEEE/CVF Conference on Computer Vision and
  Pattern Recognition}, pages 1467--1475, 2019.

\bibitem{yu2022boat}
Tan Yu, Gangming Zhao, Ping Li, and Yizhou Yu.
\newblock Boat: Bilateral local attention vision transformer.
\newblock {\em arXiv preprint arXiv:2201.13027}, 2022.

\bibitem{zhang2020cross}
Pan Zhang, Bo Zhang, Dong Chen, Lu Yuan, and Fang Wen.
\newblock Cross-domain correspondence learning for exemplar-based image
  translation.
\newblock In {\em Proceedings of the IEEE/CVF Conference on Computer Vision and
  Pattern Recognition}, pages 5143--5153, 2020.

\bibitem{zhang2018unreasonable}
Richard Zhang, Phillip Isola, Alexei~A Efros, Eli Shechtman, and Oliver Wang.
\newblock The unreasonable effectiveness of deep features as a perceptual
  metric.
\newblock In {\em Proceedings of the IEEE conference on computer vision and
  pattern recognition}, pages 586--595, 2018.

\bibitem{zhang2019multimodal}
Yulun Zhang, Chen Fang, Yilin Wang, Zhaowen Wang, Zhe Lin, Yun Fu, and Jimei
  Yang.
\newblock Multimodal style transfer via graph cuts.
\newblock In {\em Proceedings of the IEEE/CVF International Conference on
  Computer Vision}, pages 5943--5951, 2019.

\bibitem{zhang2022exact}
Yabin Zhang, Minghan Li, Ruihuang Li, Kui Jia, and Lei Zhang.
\newblock Exact feature distribution matching for arbitrary style transfer and
  domain generalization.
\newblock In {\em Proceedings of the IEEE/CVF Conference on Computer Vision and
  Pattern Recognition}, pages 8035--8045, 2022.

\bibitem{zhang2022domain}
Yuxin Zhang, Fan Tang, Weiming Dong, Haibin Huang, Chongyang Ma, Tong-Yee Lee,
  and Changsheng Xu.
\newblock Domain enhanced arbitrary image style transfer via contrastive
  learning.
\newblock In {\em ACM SIGGRAPH 2022 Conference Proceedings}, pages 1--8, 2022.

\bibitem{zhao2021improved}
Long Zhao, Zizhao Zhang, Ting Chen, Dimitris Metaxas, and Han Zhang.
\newblock Improved transformer for high-resolution gans.
\newblock {\em Advances in Neural Information Processing Systems},
  34:18367--18380, 2021.

\end{thebibliography}
}

\clearpage
\newpage
\appendix

\section{Implementation Details}

\subsection{Code}

The code will be made public soon.

\subsection{Step 1 of DA}

\begin{figure}[t]
\centering
\includegraphics[width=1.00\linewidth]{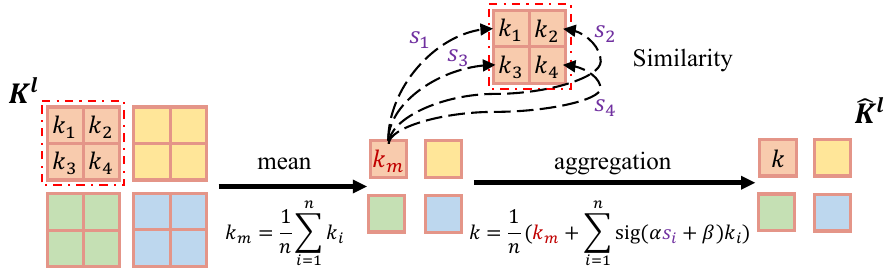}
\caption{Detailed illustration of the step 1 in DA.}
\label{fig:supp_da_step1}
\end{figure}

Fig. \ref{fig:supp_da_step1} provides a more detailed illustration of the first step in distributed attention (DA). The first step of DA is regional style aggregation. In this step, we aggregate the information of all points in a block to a point, representing the block's style information. For example, we aggregate $K_1$, $K_2$, $K_3$, and $K_4$ points in the first block of $K^l$ to a point $k$. Note that all blocks perform regional style aggregation operations in parallel, so we will finally get the regional style representations corresponding to all blocks, which form new keys $\widehat{K}^l$.

\subsection{Step 1 of PA}

\begin{figure}[t]
\centering
\includegraphics[width=1.00\linewidth]{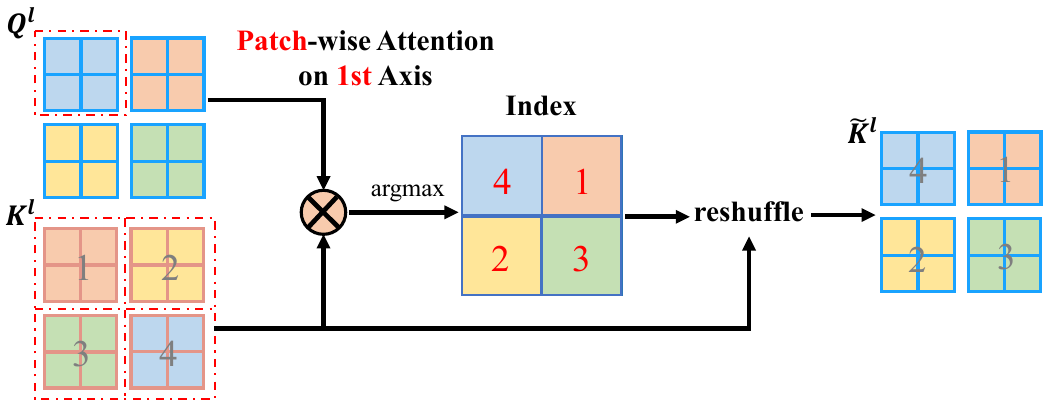}
\caption{Detailed illustration of the step 1 in PA.}
\label{fig:supp_pa_step1}
\end{figure}

Fig. \ref{fig:supp_pa_step1} gives a more detailed illustration of the first step in progressive attention (PA). The first step of PA is implemented as patch-wise attention along the first axis, which takes a block region as a token instead of a specific position. In this step, we apply argmax to the attention score to obtain the indices of the most similar coarse-grained region. For example, the $4th$ block region in $K^l$ is matched as the coarse-grained region most similar to the blue block region in $Q^l$, so the index of this region is set to 4. With the indices, we can reshuffle the tokens of $K^l$ to semantically match the spatial arrangement of the tokens of $Q^l$. The reshuffled $\widetilde{K}^l$ is further utilized in the second step of PA.

\subsection{Decoder Architecture}

The decoder implemented in this work follows the setting of \cite{liu2021adaattn}, which mirrors the encoder and takes the multi-scale transferred features as input. Full decoder configuration is shown in Table \ref{tab:supp_dec}. The decoder takes the multi-scale transferred features ${F_{cs}^l}$ as input and gradually synthesizes the final image $I_{cs}$.

\begin{table}[t]
\centering
\caption{Full configuration of the decoder.} 
\scalebox{0.95}{
\begin{tabular}{c|c|c} 
\toprule
Stage                  & Output                                                       & Architecture                               \\
\hline 
\multirow{4}{*}{$F^5$} & \multirow{4}{*}{$512 \times \frac{H}{8} \times \frac{W}{8}$} & Input $F_{cs}^5$                           \\ 
                       &                                                              & Upsample scale 2                           \\ 
                       &                                                              & Add $F_{cs}^4$                             \\ 
                       &                                                              & $3 \times 3$ Conv, 512, ReLU               \\ 
\hline
\multirow{2}{*}{$F^4$} & \multirow{2}{*}{$256 \times \frac{H}{4} \times \frac{W}{4}$} & $3 \times 3$ Conv, 256, ReLU               \\ 
                       &                                                              & Upsample scale 2                           \\
\hline
\multirow{4}{*}{$F^3$} & \multirow{4}{*}{$128 \times \frac{H}{2} \times \frac{W}{2}$} & Concatenate $F_{cs}^3$                     \\ 
                       &                                                              & ($3 \times 3$ Conv, 256, ReLU) $\times$ 3  \\ 
                       &                                                              & $3 \times 3$ Conv, 128, ReLU               \\ 
                       &                                                              & Upsample scale 2                           \\ 
\hline
\multirow{3}{*}{$F^2$} & \multirow{3}{*}{$64 \times H \times W$}                      & $3 \times 3$ Conv, 128, ReLU               \\ 
                       &                                                              & $3 \times 3$ Conv, 64, ReLU                \\
                       &                                                              & Upsample scale 2                           \\
\hline
\multirow{2}{*}{$F^1$} & \multirow{2}{*}{$3 \times H \times W$}                       & $3 \times 3$ Conv, 64, ReLU                \\ 
                       &                                                              & $3 \times 3$ Conv, 3                       \\
\bottomrule
\end{tabular}}
\label{tab:supp_dec}
\end{table}

\subsection{All-to-key Attention Algorithm}

The PyTorch code for our proposed all-to-key attention mechanism is shown in Algorithm \ref{algorithm:a2k}. The implementation is elegant with the usage of einsum notation.

\subsection{Blocking and Unblocking Algorithm}

The PyTorch code for feature blocking and unblocking operation in the all-to-key attention mechanism is shown in Algorithm \ref{algorithm:blk}

\subsection{Loss Function}

\textbf{Setting of $\lambda_1$ and $\lambda_2$} We empirically set the weight of each loss term to 10 and 1.25. Since we only have two loss terms, we can fix the weight of one loss and choose the appropriate weight according to the impact of adjusting the weight of another loss. We fix $\lambda_1$ to 10 and statistically analyze the evaluation scores of the results generated under different settings of $\lambda_2$, as shown in Table \ref{tab:supp_lamb}. As the proportion of $\lambda_2$ increases, content loss, LPIPS score decreases, and style loss increases. Therefore, there is a trade-off in style and content. We finally set $\lambda_2$ to 1.25 to render consistent style texture while maintaining semantic structure.

\begin{table}[t]
\centering
\caption{Evaluation scores of the results generated under different ratios. The best results are in \textbf{bold}.}
\scalebox{0.96}{
\begin{tabular}{c|c|c|c} 
\toprule
$\lambda_2$        & Content Loss $\downarrow$ & Style Loss $\downarrow$ & LPIPS $\downarrow$ \\
\hline  
0.5                                       & 0.72                      & \textbf{0.75}           & 0.57               \\
0.75                                      & 0.63                      & 0.82                    & 0.55               \\
1                                         & 0.58                      & 0.98                    & 0.54               \\
1.25                                      & 0.55                      & 1.04                    & 0.53               \\
1.5                                       & 0.52                      & 1.20                    & 0.52               \\
2                                         & \textbf{0.49}             & 1.38                    & \textbf{0.49}      \\
\bottomrule
\end{tabular}}
\label{tab:supp_lamb}
\end{table}


\section{More Results}

\subsection{All-to-key Attention VS. All-to-all Attention}

\begin{figure*}[t]
\centering
\includegraphics[width=1.00\linewidth]{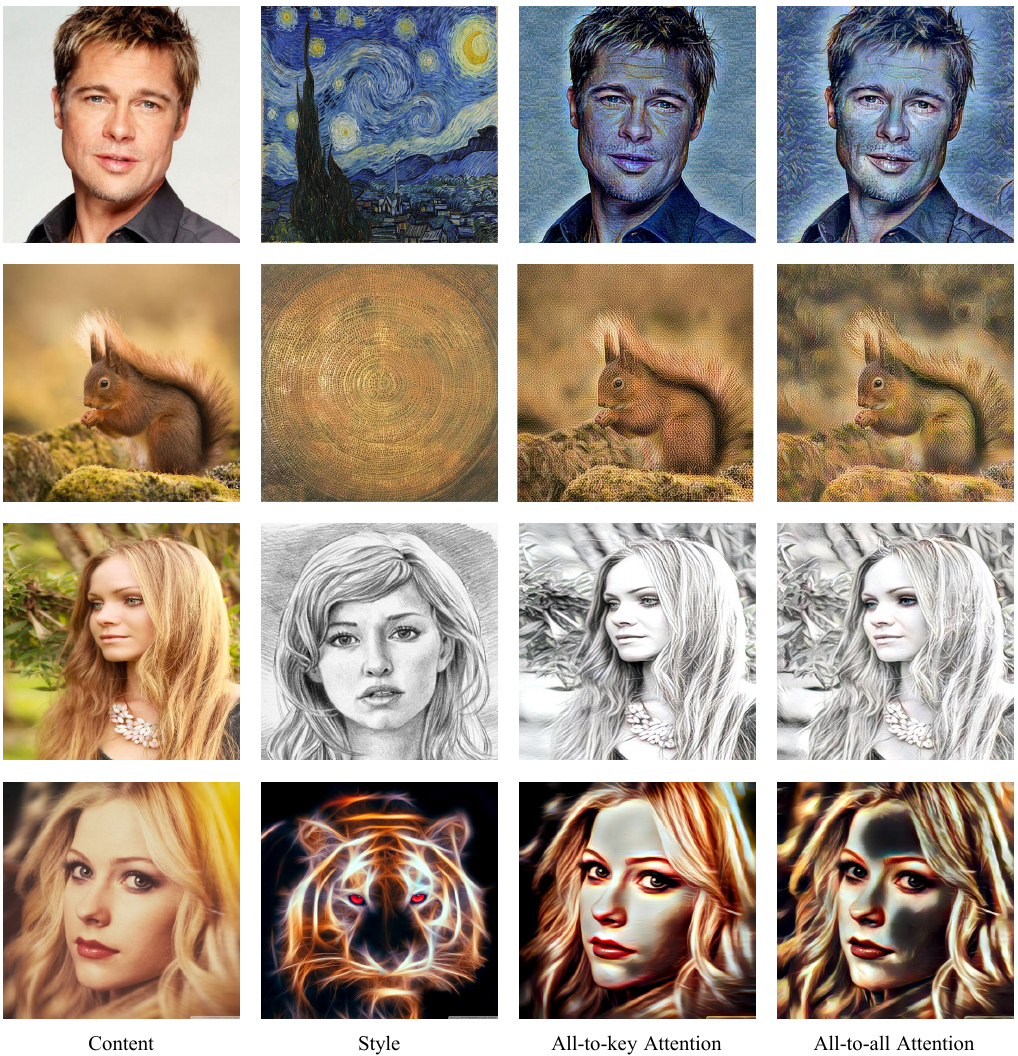}
\caption{More visual comparisons between our proposed all-to-key attention and all-to-all attention.}
\label{fig:supp_abla}
\end{figure*}

To further illustrate the superiority of our proposed all-to-key attention to all-to-all attention in producing high-quality stylized images, we provide more visual comparisons in Figure \ref{fig:supp_abla}. By replacing all-to-key attention in our full model with all-to-all attention, the visual quality of the stylized images decreases significantly, affected by distorted style patterns.


\subsection{Arbitrary Style Transfer}

\begin{figure*}[t]
\centering
\includegraphics[width=1.00\linewidth]{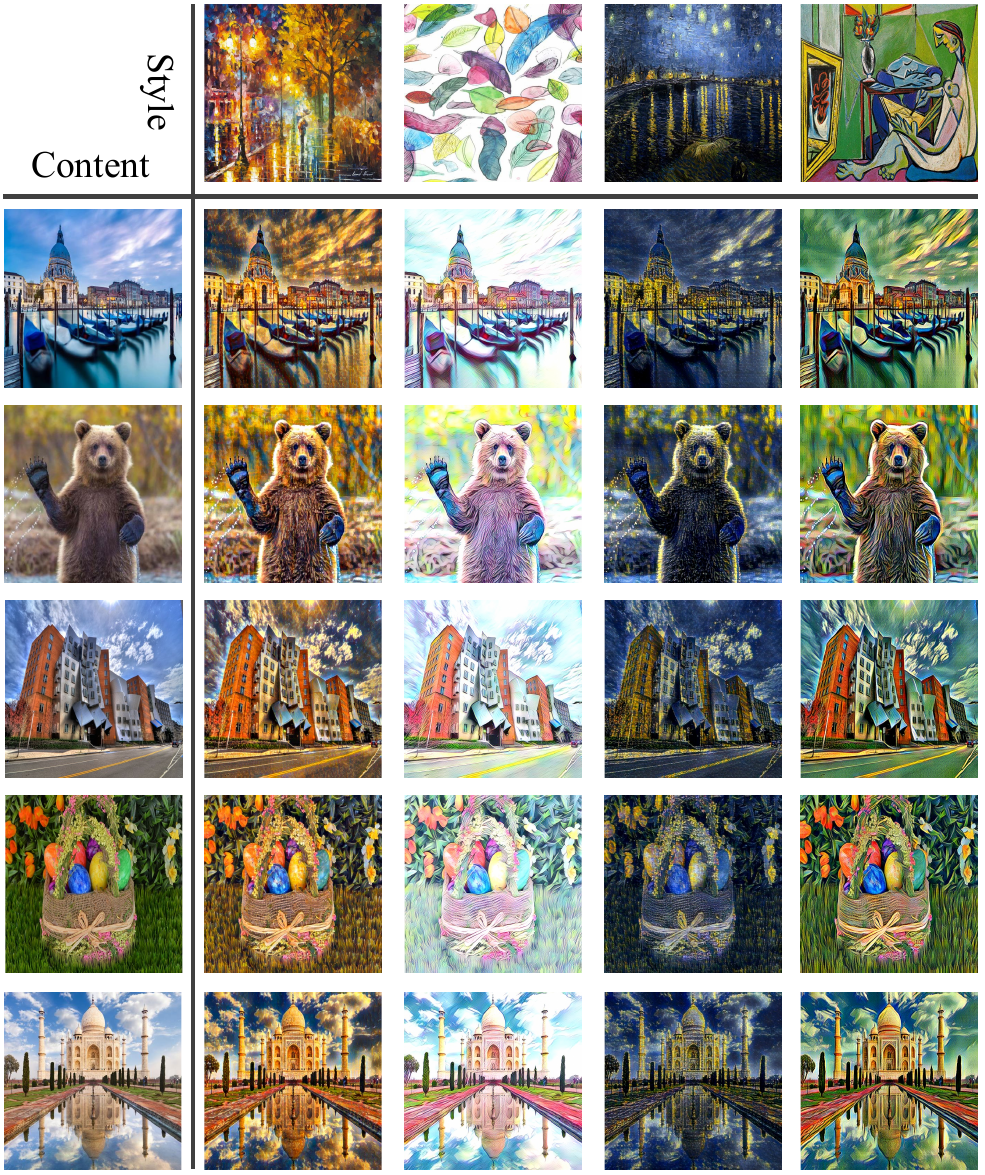}
\caption{More image stylization results.}
\label{fig:supp_res1}
\end{figure*}

\begin{figure*}[t]
\centering
\includegraphics[width=1.00\linewidth]{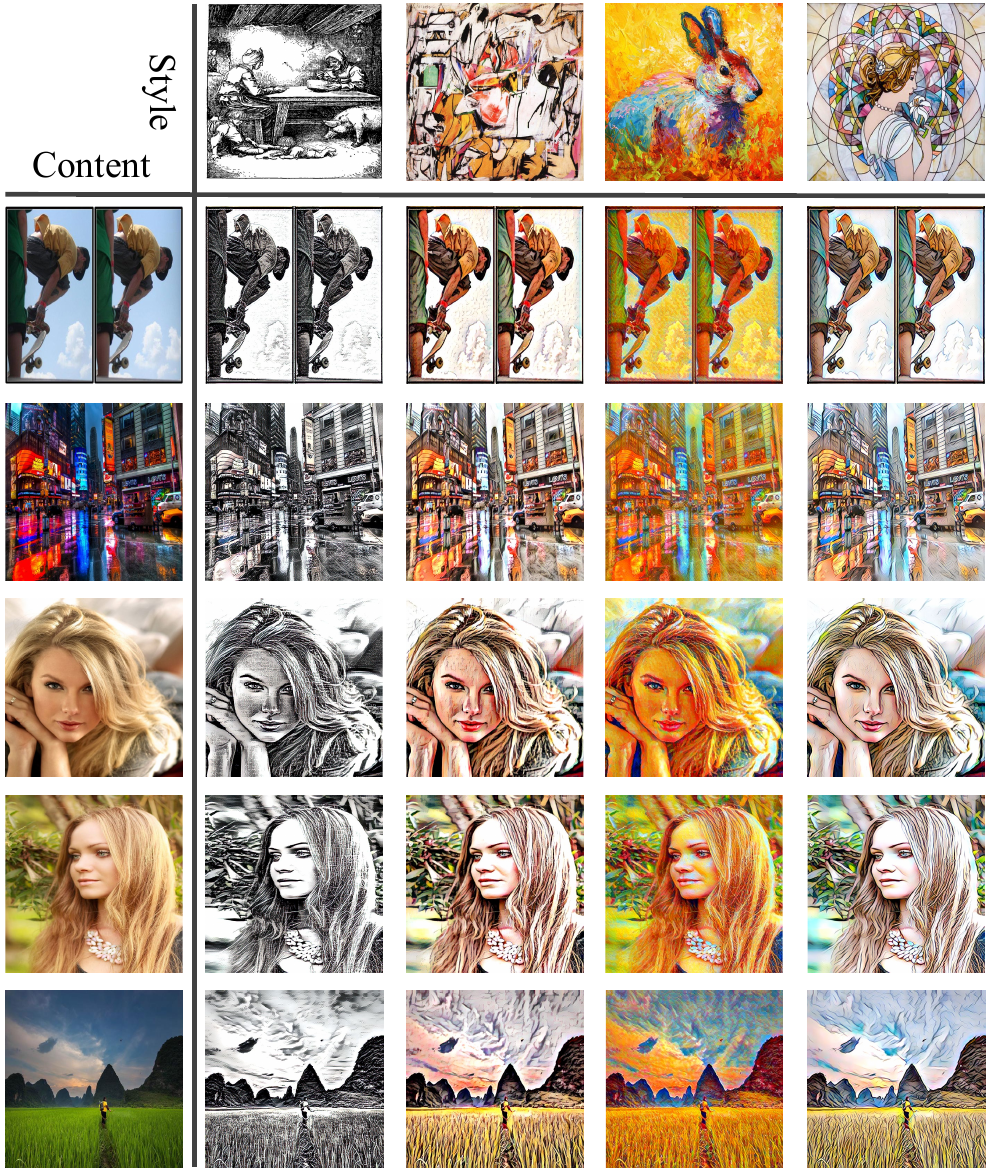}
\caption{More image stylization results.}
\label{fig:supp_res2}
\end{figure*}

To further demonstrate the effectiveness and robustness of our proposed StyA2K on arbitrary style transfer, we provide more stylization results of pair-wise combinations between 10 content images and 8 style images (total 80 stylized images) in Figure \ref{fig:supp_res1} and Figure \ref{fig:supp_res2}. Our method can faithfully generate visually appealing results with consistent style textures.

\begin{algorithm*}[t]  
\caption{Pytorch code implementing all-to-key attention.}  
\begin{lstlisting}
class A2K(nn.Module):
    '''All-to-key Attention'''
def __init__(self, in_dim):
    super().__init__()
    self.Dq = nn.Conv2d(in_dim, in_dim, (1, 1))
    self.Dk = nn.Conv2d(in_dim, in_dim, (1, 1))
    self.Dv = nn.Conv2d(in_dim, in_dim, (1, 1))
    self.Pq = nn.Conv2d(in_dim, in_dim, (1, 1))
    self.Pk = nn.Conv2d(in_dim, in_dim, (1, 1))
    self.Pv = nn.Conv2d(in_dim, in_dim, (1, 1))
    self.sim_alpha = nn.Parameter(torch.ones(1), require_grad=True)
    self.sim_beta = nn.Parameter(torch.zeros(1), require_grad=True)
    self.fusion_D = nn.Conv2d(in_dim, in_dim, (1, 1))
    self.fusion_P = nn.Conv2d(in_dim, in_dim, (1, 1))

def forward(self, content, style):
    # Get Q K V
    DA_q = block(self.Dq(mean_variance_norm(content)), p_size, stride)
    DA_k = block(self.Dk(mean_variance_norm(style)), p_size, stride)
    DA_v = block(self.Dv((style)),p_size,stride)
    PA_q = block(self.Pq(mean_variance_norm(content)), p_size, stride)
    PA_k = block(self.Pk(mean_variance_norm(style)), p_size, stride)
    PA_v = block(self.Pv(style), p_size, stride)
    # Distributed Attention Step 1
    DA_k_m = torch.mean(DA_k, dim=-1)
    DA_v_m = torch.mean(DA_v, dim=-1)
    dis = torch.einsum("bhcx,bhcxy->bhxy", DA_k_m, DA_k)
    sim = torch.sigmoid(self.sim_beta+self.sim_alpha*dis)
    DA_k_a = (torch.einsum("bhxy,bhcxy->bhcx", sim, DA_k) + DA_k_m) / p_size
    DA_v_a = (torch.einsum("bhxy,bhcxy->bhcx", sim, DA_v) + DA_v_m) / p_size
    # Distributed Attention Step 2
    logits = torch.einsum("bhcxy,bhczy->bhyxz", DA_q, DA_k_a)
    scores =  softmax(logits)
    DA = torch.einsum("bhyxz,bhvzy->bhcxy", scores, DA_v_a)
    DA_unblock = unblock(DA)
    # Progressive Attention Step 1
    PA1_logits = torch.einsum("bhcxy,bhczy->bhxz", PA_q, PA_k)
    index = torch.argmax(PA1_logits, dim = -1).expand_as(PA_k)  
    PA_k_reshuffle = torch.gather(PA_k, -2, index)
    PA_V_reshuffle = torch.gather(PA_v, -2, index)
    # Progressive Attention Step 2
    logits2 = torch.einsum("bhcxy,bhcxz->bhxyz", PA_q, PA_k_reshuffle)
    scores2 = softmax(logits2) 
    PA = torch.einsum("bhxyz,bhcxz->bhcxy", scores2, PA_V_reshuffle)
    PA_unblock = unblock(O2)   
    # Feature Transformation
    O_DA = self.fusion_D(DA_unblock)
    O_PA = self.fusion_P(PA_unblock)
    O = O_DA + O_PA
    out = O + Content   
    return out        
\end{lstlisting}
\label{algorithm:a2k}  
\end{algorithm*}

\begin{algorithm*}[t]  
\caption{Pytorch code implementing feature blocking and unblocking.}  
\begin{lstlisting}
def block(X, patch_size, stride):
    '''feature blocking.
    Args:
        X: a tensor with shape [b, c, h, w], where b is batch size, c is the 
        channel dimension, h is feature height, and w is feature width.
        patch_size: an integer for the patch (block) size
        stride: the parameter of torch.nn.functional.unfold
    Returns:
        Y: a tensor with shape [b, c, n, r], where n is patch sequence length
        and r is the patch size. 
    '''
    b, c, h, w = X.shape
    r = int(patch_size**2)
    Y = torch.nn.functional.unfold(X, kernel_size=patch_size, stride=stride)
    Y = Y.view(b, c, r, -1).permute(0, 1, 3, 2)
    return Y

def unblock(X, patch_size, stride, h):
    '''feature unblocking.
    Args:
        X: a tensor with shape [b, c, n, r], where b is batch size, c is 
        channel dimension, n is patch sequence length, r is the patch size.
        patch_size: an integer for the patch (block) size
        stride: the parameter of torch.nn.functional.unfold
        h: the output height of the feature 
    Returns:
        Y: a tensor with shape [b, c, h, w], where h is is feature height 
        and w is feature width. 
    '''
    b, c, n, r = X.shape
    X = X.permute(0, 2, 1, 3)
    X = X.contiguous().view(b, n, -1).permute(0, 2, 1)
    Y = torch.nn.functional.unfold(X, h, kernel_size=patch_size, stride=stride)
    return Y
\end{lstlisting}
\label{algorithm:blk}  
\end{algorithm*}

\end{document}